\documentclass{article}

\usepackage{microtype}
\usepackage{graphicx}
\usepackage{subfigure}
\usepackage{amsmath}
\usepackage{amsfonts}
\usepackage{enumitem}
\usepackage{amsthm}
\newtheorem{theorem}{Theorem}
\usepackage{xcolor}
\usepackage{natbib}
\usepackage{textcomp, libertine}
\usepackage{booktabs} 
\usepackage{hyperref}

\usepackage[accepted]{icml2019}

\icmltitlerunning{Compositional Fairness Constraints for Graph Embeddings}

\newcommand{\G}{\mathcal{G}}
\newcommand{\V}{\mathcal{V}}
\newcommand{\E}{\mathcal{E}}
\newcommand{\T}{\mathcal{T}}
\newcommand{\A}{\mathcal{A}}
\newcommand{\R}{\mathcal{R}}
\newcommand{\enc}{\textsc{enc}}
\newcommand{\compenc}{\textsc{c-enc}}

\newcommand{\xhdr}[1]{{\noindent\bfseries #1}.}
\newcommand{\cut}[1]{}
\newcommand{\CITE}{\textcolor{red}{CITE}}

\newcommand{\mb}{\mathbf}
\newcommand{\removelatexerror}{\let\@latex@error\@gobble}
\begin{document}

\twocolumn[
\icmltitle{Compositional Fairness Constraints for Graph Embeddings}



\icmlsetsymbol{equal}{*}

\begin{icmlauthorlist}
\icmlauthor{Avishek Joey Bose}{mcgill,mila}
\icmlauthor{William L. Hamilton}{mcgill,mila,fair}
\end{icmlauthorlist}

\icmlaffiliation{mcgill}{McGill University}
\icmlaffiliation{mila}{Mila}
\icmlaffiliation{fair}{Facebook AI Research}

\icmlcorrespondingauthor{Avishek Joey Bose}{joey.bose@mail.mcgill.ca}

\icmlkeywords{Machine Learning, ICML}

\vskip 0.3in
]



\printAffiliationsAndNotice{}  
\pdfoutput=1
\begin{abstract}
 Learning high-quality node embeddings is an important building block for machine learning models that operate on graph data, such as social networks and recommender systems. However, existing graph embedding techniques are unable to cope with fairness constraints, e.g., ensuring that the learned representations do not correlate with certain attributes, such as age or gender. Here, we introduce an adversarial framework to enforce fairness constraints on graph embeddings. Our approach is {\em compositional}---meaning that it can flexibly accommodate different combinations of fairness constraints during inference. 
 For instance, in the context of social recommendations, our framework would allow one user to request that their recommendations are invariant to both their age and gender, while also allowing another user to request invariance to just their age. 
 Experiments on standard knowledge graph and recommender system benchmarks highlight the utility of our proposed framework. 
\end{abstract}
\section{Introduction}
Learning low-dimensional embeddings of the nodes in a graph is a fundamental technique underlying state-of-the-art approaches to link prediction and recommender systems \cite{hamilton2017representation}. 
However, in many applications---especially those involving social graphs---it is desirable to exercise control over the information contained within learned node embeddings. 
For instance, we may want to ensure that recommendations are fair or balanced with respect to certain attributes (e.g., that they do not depend on a user's race or gender) or we may want to ensure privacy by not exposing certain attributes through learned node representations. 
In this work we investigate the feasibility of enforcing such {\em invariance constraints} on (social)  graph embeddings.

\begin{figure}[t!]
    \centering
    \includegraphics[width=1\linewidth]{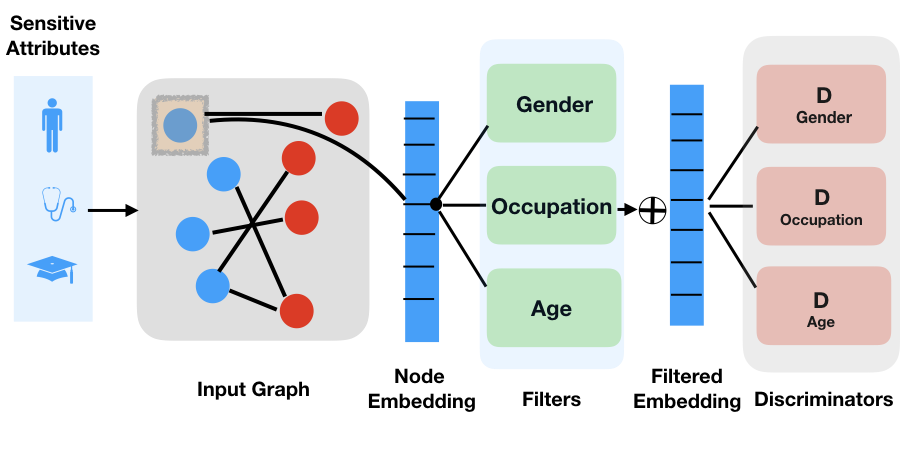}
    \vspace{-5mm}
    \caption{\textbf{Overview of our approach:} Our goal is to generate graph embeddings that are invariant to particular sensitive attributes (e.g., age or gender). We train a set of ``filters'' to prevent adversarial discriminators from classifying the sensitive information from the filtered embeddings. After training, these filters can be composed together in different combinations, allowing the flexible generation of embeddings that are invariant w.r.t. any subset of the sensitive attributes.}
    \label{arch}
    \vspace{-5mm}
\end{figure}

While enforcing invariance constraints on general classification models \cite{chouldechova2017fair,gajane2017formalizing,kamishima2012fairness} and collaborative filtering algorithms \cite{yao2017new} has received considerable attention in recent years, these techniques have yet to be considered within the context of graph embeddings---a setting that introduces particular challenges due to the non-i.i.d. and non-Euclidean nature of relational, graph data. 

Moreover, in the case of social graphs and large-scale recommender systems, it is often the case that there are many  possible sensitive attributes that we {\em may} want to enforce invariance constraints over.
Previous work on enforcing invariance (or ``fairness'') in social applications has generally focused on situations that involve one sensitive attribute (e.g., age in the context of credit or loan decisions; \citet{zemel2013learning}), but in the context of social graph embeddings there can be an extremely large number of possible sensitive attributes. In fact, in extreme settings we may even want to be fair with respect to the existence of individual edges.
For instance, a user on a social networking platform might want that platform's recommender system to ignore the fact that they are friends with a certain other user, or that they engaged with a particular piece of content. 

\xhdr{Our contributions}
We introduce an adversarial framework to enforce {\em compositional} fairness constraints on graph embeddings for multiple sensitive attributes.
The insight behind our approach is that we learn a set of {\em adversarial filters} that remove information about particular sensitive attributes.
Importantly, each of these learned filters can be {\em optionally} applied after training, so the model can flexibly generate embeddings that are invariant with respect to different combinations of sensitive attributes. As the space of possible combinations of sensitive attributes can be combinatorially large, we demonstrate that our compositional strategy can generate invariant embeddings even on unseen combinations at test time. Our contribution is at the intersection of research on (social) graph embedding and algorithmic fairness.
We build upon the success of recent adversarial approaches to fairness \cite{edwards2015censoring}, disentanglement \cite{mathieu2016disentangling}, and transfer learning \cite{madras2018learning}--extending these approaches to the domain of graph representation learning and introducing new algorithmic techniques to accommodate compositional constraints during inference. 

\section{Related Work}
We now briefly highlight core related work on (social) graph embeddings and algorithmic fairness, which our research builds upon.

\subsection{Graph Embedding}
At the core of our proposed methodology is the notion of learning low-dimensional embeddings of graph-structured data, especially social data.
Graph embedding techniques have a long history in the social sciences, with connections to early research on ``sociograms'' (small hand-constructed social networks) and latent variable models of social interactions \cite{faust1988comparison,majone1972social}.
In more recent years, the task of embedding graph-structured data has received increasing attention from the machine learning and data mining communities \cite{cai2018comprehensive,hamilton2017representation}. 
Generally, the goal of these works is to map graph nodes to low-dimensional vector embeddings, such that the original graph can be reconstructed from these embeddings.
Traditional approaches to this problem include Laplacian eigenmaps \cite{belkin2002laplacian} and matrix factorization techniques \cite{ng2001spectral}, with recent years witnessing a surge in methods that rely on random-walk based objectives \cite{grover2016node2vec,perozzi2014deepwalk}, deep autoencoders \cite{wang2016structural}, and graph neural networks \cite{hamilton2017inductive,kipf2016variational}. 

Learned graph embeddings can be used for a wide variety of tasks, including node classification, relation prediction, and clustering \cite{hamilton2017representation}.
Here, we focus on the relation prediction task, i.e., using the learned representations to predict previously unobserved relationships between the input nodes.
The relation prediction task is exceptionally general---for example, it generalizes basic recommender systems, knowledge base completion, and even node classification (\textit{ibid.}).

\subsection{Algorithmic Fairness}\label{sec:relatedfair}
Unlike previous research on graph embedding, in this work we focus on the challenge of enforcing fairnes or invariance constraints on the learned representations. 
Recent work on fairness in machine learning, including work on fairness in collaborative filtering, involves making predictions that are balanced or invariant with respect to certain sensitive variables (e.g., age or gender) \cite{chouldechova2017fair,gajane2017formalizing,kamishima2012fairness,madras2018learning,zemel2013learning,yao2017new}. 
Formally, in the standard ``fair classification'' setting we consider a data point $\mb{x} \in \mathbb{R}^n$, its class label $y \in \mathcal{Y}$, and a binary sensitive attribute $a \in \{0,1\}$ (e.g., indicating gender). 
The high-level goal is then to train a model to predict $y$ from $\mb{x}$, while making this prediction invariant or fair with respect to $a$ \cite{madras2018learning}.
There are many specific definitions of fairness, such as whether fairness refers to parity or satisfying certain preferences (see \cite{gajane2017formalizing} for a detailed discussion).
In the context of fair machine learning, our core contribution is motivating, implementing, and evaluating an approach to enforce fairness within the context of graph embeddings.
There are a number of complications introduced by this setting---for instance, rather than one classification task, we instead have thousands or even millions of interdependent edge relationships. 

\section{Preliminaries}
We consider the general case of embedding a heterogeneous or multi-relational (social) graph $\G = (\V, \E)$, which consists of a set of directed edge triples $e=\langle u, r, v \rangle \in \E$, where $u, v \in \V$ are nodes and $r \in \R$ is a relation type.
We further assume that each node is of a particular type, $\T \subseteq \V$, and that relations may have constraints regarding the types of nodes that they can connect. 

\xhdr{Relation Prediction}
The general {\em relation prediction} task on such a  graph is as follows.
Let $\E_{\textrm{train}}\subset \E$ denote a set of observed {\em training} edges and let $\bar{\E} = \{\langle v_i, r, v_j \rangle : v_i, v_j \in \V, r \in R\} \setminus \E$ denote the set of {\em negative} edges that are not present in the true graph $\G$.
Given $\E_{\textrm{train}}$, we aim to learn a scoring function $s$ such that
\begin{equation}\label{eq:score}
    s(e) > s(e'), \forall e \in \E, e' \in \bar{\E}.
\end{equation}
In other words, the learned scoring function should ideally score any true edge higher than any negative edge. 

\xhdr{Embedding-based Models}
In the context of graph embeddings, we aim to solve this relation prediction task by learning a function $\enc : \V \mapsto \mathbb{R}^d$ that maps each node $v \in \V$ to an embedding $\mb{z}_v = \enc(v)$.
In this case, the signature of the score function becomes $s : \mathbb{R}^d \times \R \times \mathbb{R}^d \mapsto \mathbb{R}$, i.e., it takes two node embeddings $\mb{z}_u, \mb{z}_v \in \mathbb{R}^d$ and a relation $r \in \R$ and scores the likelihood that the edge $e=<u, r, v>$ exists in the graph. 
Generally, the intuition in embedding-based approaches is that the distance between two node embeddings should encode the likelihood that there is an edge between the nodes.
\cut{
For instance, a concrete example of this embedding-based approach is the TransE \cite{bordes2013translating} model, where each relation $r \in \R$ is associated with an embedding $\mb{r} \in \mathbb{R}^{d}$ and the scoring function is defined as
\begin{align}
    s(\mb{z}_u, r, \mb{z}_v) &=  - \|\mb{z}_u^\top + \mb{r} - \mb{z}_v\|_2,
\end{align}
i.e., the likelihood of a relation holding between two nodes is proportional to their distance, after translating one of the node embeddings by a learned relation embedding. 
TransE is just one of many possible instantiations of this general approach, and many scoring functions have been proposed in recent years---including RESCAL \cite{nickel2011three}, ComplexE \cite{trouillon2016complex} and TransD \cite{ji2015knowledge}.
}
Following standard practice, we consider the optimization of these scoring functions using contrastive learning methods that make use of a corruption distribution such as {\em noise contrastive estimation} \cite{dyer2014notes,mnih2012fast} and similar variants \cite{bose2018adversarial}, where the loss over a batch of edges 
$\E_{\textrm{batch}} \subseteq \E_{\textrm{train}}$ is given by:
\begin{equation}\label{eq:genericloss}
    \sum_{e \in \E_{\textrm{edge}}} L_{\textrm{edge}}(s(e), s(e^{-}_1), ..., s(e^{-}_m)),
\end{equation}
where $L_{\textrm{edge}}$ is a per-edge loss function and $e^{-}_1, ..., e^{-}_m \in \bar{\E}$ are ``negative samples'', i.e., randomly sampled edges that do not exist in the graph.
Loss functions of this form generally attempt to maximize the likelihood of true edges compared to the negative samples.

\xhdr{Fairness}
In order to incorporate the notion of fairness into the graph embedding setup, we assume that for exactly one node type $\T^*$, all nodes of this type, i.e. all $u \in \T^*$, have $K$ categorical sensitive attributes, $a^k_u \in \A_k, k=1...,K$, and for simplicity, we assume that there are no other features or attributes associated with the nodes and edges in the graph.\footnote{Though this assumption can easily be relaxed.}
The challenge in enforcing fairness is thus to ensure that the learned node embeddings, $\mb{z}_u$, are not biased or unfair with respect to these sensitive attributes---a point which we formalize in the next section. 
\cut{For example, in a simple social recommendation setting, we may have one type of node, $\T = \{\texttt{users}\}$ and one type of relation, $\mathcal{R} = \{\texttt{friends}\}$.
In the basic graph embedding setting, our goal would be to learn embeddings of the users, which we could use to make friend recommendations. 
However, in the case of enforcing fairness, we might further assume that all the users have a categorical attributes indicating their age and gender, and our goal would to be recommend friends in such a way that does not depend on these attributes. }

\section{Invariant Graph Embeddings}
We first motivate and argue in favor of a particular form of ``fairness'' (or rather {\em invariance}) within the context of graph embeddings.
Following this, we outline our compositional and adversarial approach for enforcing these invariance constraints on graph embeddings. 

\subsection{Pragmatic Fairness as Invariance}

In this paper we consider a simple, user-centric formulation of fairness within the context of social graph embeddings. \cut{
Thus, rather than developing yet another new metric---or diving into thorny debates regarding the pros and cons of different formulations---in this work we instead consider a simple, user-centric formulation within the context of social graph embeddings.}
Using gender as an example of a sensitive attribute and movie recommendation as an example relation prediction task, our approach is guided by the following question: {\em If one gives a user a button that says ``Please ignore my gender when recommending movies'', what does a user expect from the system after this button is pressed?}
Here, we accept it as non-controversial that the expectation from the user is that recommendation does not depend in any way on their gender, i.e., that the recommendation would be the same regardless of their gender.
Formally, given a user $u$, this expectation amounts to an assumption of independence, 
\begin{equation}\label{eq:edgeind}
  s(e) \perp a_u \qquad \forall v \in \V, r \in \R
\end{equation}
between the recommendation---i.e., the score of the edge, $s(e) = s(\langle \mb{z}_u, r, \mb{z}_v\rangle)$---and the sensitive attribute $a_u$.

One issue in directly enforcing Equation \eqref{eq:edgeind} is that there are many (potentially millions) of possible edges that we might want to score for every node $u \in \T^*$, making it intractable to enforce independence on each of these decisions individually.
However, if we assume that the score function $s(\langle \mb{z}_u, r, \mb{z}_v \rangle)$ depends on $u$ only through $u's$'s embedding, $\mb{z}_u$, then we can guarantee the independence in Equation \eqref{eq:edgeind} for all edge predictions by enforcing what we call {\em representational invariance}:
\begin{equation}\label{eq:repinvar}
    \mb{z}_u \perp a_u, \:\: \forall u \in \V.
\end{equation}
In other words, we require that the mutual information $I(\mb{z}_u, a_u)$ is $0$.

Generalizing to the setting of multiple sensitive attributes, for a given set of sensitive attributes $S \subseteq \{1, ..., K\}$, we would require that
\begin{equation}\label{eq:multirepinvar}
     I(\mb{z}_u, a^k_u) =0, k\in S, \forall u \in \V,
\end{equation}
which amounts to the assumption of $S$ independent invariance constraints on the $S$ distinct sensitive attributes.\footnote{Note that this does not necessarily imply ``subgroup fairness'' on the joint distribution \cite{kearns2017preventing}.}
Importantly, {\em we assume that the set $S$ is not fixed} (e.g., different users might request different invariance constraints). 
In the language of algorithmic fairness, the representational invariance of Equation \eqref{eq:multirepinvar} implies that traditional demographic parity constraints are satisfied on the sensitive attributes and recommendations.

\subsection{Model Definition}

In this work, we enforce representational invariance constraints on the node embeddings (Equation \ref{eq:multirepinvar}) by introducing an adversarial loss and a technique to ``filter'' the embeddings generated by the $\enc$ function.
Note, again, that a unique challenge here is that $S$---the set of sensitive attributes we want to be invariant with respect to---is not  fixed across nodes; i.e., we may want to enforce invariance on different sets of sensitive attributes for different nodes. 

Note also that the framework presented in this section is quite general and can function with arbitrary combinations of base node embedding functions $\enc$ and edge-prediction losses $L_{\textrm{edge}}$ (see Equation \ref{eq:genericloss}).
We discuss three concrete instantiations of this framework in Section \ref{sec:experiments}.

\xhdr{Compositional Encoder}
The first important insight in our model is generalizing the $\enc$ embedding function to optionally ``filter'' out the information about certain sensitive attributes. 
In particular, for every sensitive attribute $k \in \{1, ..., K\}$ we define a filter function $f_k : \R^d \mapsto \R^d$ that is trained to remove the information about the $k$th sensitive attribute.
If we want the node embedding to be invariant w.r.t. some set of sensitive attributes $S \subseteq \{1,...,K\}$, we then generate its embedding by ``composing'' the output of the $|S|$ filtered embeddings using a {\em compositional encoder}:
\begin{equation}\label{eq:compenc}
    \compenc(u, S) = \frac{1}{|S|}\sum_{k \in S}f_k(\enc(u))
\end{equation}
To train $\compenc(u, S)$ we sample a binary mask to determine the set $S$ at every iteration. 
In this work, we sample the binary mask as a sequence of $k$ independent Bernoulli draws with a common fixed probability $p=0.5$; however, other application-specific distributions (e.g., incorporating dependencies between the attributes) could be employed.
Sampling random binary masks forces the model to produce invariant embeddings for different combinations of sensitive attributes during training with the hope of generalizing to unseen combinations during inference time --- a phenomena that we empirically validate in Section \ref{subsubsec:comp_generalizability}.

\xhdr{Adversarial Loss}
To train the compositional encoder, we employ an adversarial regularizer. 
For each sensitive attribute $k \in K$, we define a discriminator $D_k : \mathbb{R}^{d} \times \A_k \mapsto [0,1]$, which attempts to predict the $k$th sensitive attribute from the node embeddings. 
Assuming we are given an edge-prediction loss function $L_\textrm{edge}$ (as in Equation \ref{eq:genericloss}), we can then define our new adversarially regularized per-edge loss as
\begin{align}\label{eq:loss}
    L(e) = &L_{\textrm{edge}}(s(e), s(e^{-}_1), ..., s(e^{-}_m)) \nonumber \\ &+ \lambda \sum_{k \in S}\sum_{a^k \in \A_k}\log(D_k(\compenc(u, S), a^k)),
\end{align}
where $\lambda$ is a hyperparameter controlling the strength of the adversarial regularization.
To optimize this loss in a minibatch setting, we alternate between two types of stochastic gradient descent updates: (1) $T$ minibatch updates minimizing $L(e)$ with respect to $\compenc$ (with all the $D_k$ fixed), and (2) $T'$ minibatch updates minimizing $-L(e)$ with respect to $D_k, k=1...,K$ (with $\compenc$ fixed).

\xhdr{Theoretical Considerations}
For clarity and simplicity, we consider the case of a single binary sensitive attribute, with the theoretical intuitions naturally generalizing to the multi-attribute and multi-class settings.
Assuming a single binary sensitive attribute $a_k$, by simple application of Proposition 2 in \citet{goodfellow2014generative}, we have:\footnote{Theorem 1 holds as a consequence of Proposition 2 in \citet{goodfellow2014generative} if we simply replace the task of distinguishing real/fake data by classifying a binary sensitive attribute. }
\begin{theorem}
If $\compenc$ and $D_k$ have enough capacity, $T'$ is large enough so that $D_k$
is allowed to reach its optimum on $-L(e)$ (with $\compenc$ fixed), and $\compenc$ is optimized according to $L(e)$ (with $D$ fixed), then $I(\mb{z}_u, a_u) \rightarrow 0, \forall u \in \T^*$ as $\lambda \rightarrow \infty$. 
\end{theorem}

That is, if we increase the weight of the adversarial regularizer to infinity, the equilibrium of the minimax game in Equation \eqref{eq:loss} occurs when there is zero mutual information between the sensitive attribute and the embeddings. 
Of course, as $\lambda \rightarrow \infty$ trivial solutions to this game exist (e.g., $\compenc$ simply outputting a constant value) and in practice setting $\lambda < \infty$ leads to a tradeoff between performance on edge prediction and representational invariance.

\section{Experiments}\label{sec:experiments}

We investigated the impact of enforcing invariance on graph embeddings using three  datasets: Freebase15k-237\footnote{\tiny{\url{www.microsoft.com/en-us/download/details.aspx?id=52312}}}, MovieLens-1M\footnote{\tiny{\url{grouplens.org/datasets/movielens/1m/}}}, and an edge-prediction dataset derived from Reddit.\footnote{\tiny Using data from \url{ https://pushshift.io}, a previously existing dataset collected by Jason Baumgartner. The authors and their institutions were not involved in the data collection.} The dataset statistics are given in Table \ref{dataset_stats}. Our experimental setup closely mirrors that of \cite{madras2018learning} where we jointly train the main model with adversaries, but when testing invariance, we train a \textit{new} classifier (with the same capacity as the discriminator) to predict the senstive attributes from the learned embeddings.

The goal of our experiments was to answer three questions:
\begin{enumerate}[label={(\bf Q\arabic*)}, topsep=0pt, parsep=0pt, leftmargin=25pt, itemsep=2pt]
    \item \textbf{The invariance-accuracy tradeoff}. What is the tradeoff between enforcing invariance and accuracy on the main edge prediction task?
    \item \textbf{The impact of compositionality.} How does the performance of a compositional approach, which jointly enforces fairness over a set of sensitive attributes, compare to a more traditional model that only enforces fairness on a single attribute?
    \item \textbf{Invariance on unseen combinations.} In settings with many sensitive attributes, is our approach able to enforce invariance even on combinations of sensitive attributes that it never saw during training? 
\end{enumerate}
Throughout these experiments, we rely on two  baselines: First, we compare against baselines that do not include any invariance constraints, i.e., models with $\lambda=0$.
Second, we compare against a non-compositional adversarial approach where we separately train $K$ distinct encoders and $K$ distinct adversaries for each of the $K$ sensitive attributes in the data. 
This non-compositional adversary is essentially an extension of \citet{edwards2015censoring}'s approach to the graph embedding domain. 

\subsection{Setup and Datasets}

\begin{table*}[t]
\caption{Statistics for the three datasets, including the total number of nodes $(|\V|$) and number of nodes with sensitive attributes $|\T^*|$, the number of sensitive attributes and their types and the total number of edges in the graph.}
\label{datasetstats-table}
\begin{center}
\begin{small}
\begin{sc}
\begin{tabular}{lcccccr}
\toprule
Dataset & $|\V|$ & $|\T^*|$ &\multicolumn{1}{p{1.5cm}}{\centering \#Sensitive \\ Attributes} & Edges & \multicolumn{1}{p{1.5cm}}{\centering Binary \\ Attributes?} & \multicolumn{1}{p{1.5cm}}{\centering Multiclass \\ Attributes?} \\
\midrule
FB15k-237    & 14,940 & 14,940 & 3 & 168,618 &$\surd$ & $\times$ \\
MovieLens1M & 9,940& 6,040& 3& 1,000,209&$\surd$ &$\surd$\\
Reddit Comments    & 385,735 & 366,797& 10 &7,255,096 &$\surd$ & $\times$ \\
\bottomrule
\end{tabular}
\end{sc}
\end{small}
\end{center}
\vskip -0.1in
\label{dataset_stats}
\end{table*}

Before describing our experimental results, we first outline some important properties of the datasets we used, as well as the specific encoders and edge-prediction loss functions used.  

In all experiments, we used multi-layer perceptrons (MLPs) with leaky ReLU activation functions \cite{xu2015empirical} as the discriminators $D_k$ and filters $f_k$.
The Appendix contains details on the exact hyperparameters (e.g., number of layers and sizes) used for all the different experiments, as well as details on the training procedures (e.g., number of epochs and data splits).
Code to reproduce our results is available at:
\url{https://github.com/joeybose/Flexible-Fairness-Constraints}.

\subsubsection*{Freebase15k-237}
\vspace{-2mm}
Freebase 15k-237 is a standard benchmark used for knowledge base completion \cite{toutanova2015representing}.
In this work, we use Freebase 15k-237 as a semi-synthetic testbed to evaluate the impact of adversarial regularization. 
Taking the entity attribute labels from \citet{moon2017learning}, we used the $3$-most common attribute labels (e.g., /award/award\_nominee) as ``sensitive'' attributes.
The goal in this dataset is to perform the standard knowledge base completion task, while having the entity embeddings be invariant with respect to these ``sensitive'' attribute labels.
While synthetic, this dataset provides a useful reference point due to its popularity in the graph embedding literature. 

For our encoder and edge-prediction loss function, we follow \citet{ji2015knowledge}'s TransD approach, since we found this approach gave significant performance boosts compared to simpler models (e.g., TransE). 
In this model, the encoding of a node/entity depends on the edge relation being predicted, as well as on whether the entity is the head or tail in a relation (i.e., the edge direction matters).
In particular, the embedding of the head node (i.e., the source node) in an edge relation is given by:
\begin{equation}
    \enc(u, \langle u, r, v \rangle) = (\mb{r}_p\mb{u}_p^\top + \mb{I}^{d \times d})\mb{u},
\end{equation}
where $\mb{u}, \mb{u}_p, \mb{r}_p \in \mathbb{R}^d$ are trainable embedding parameters and $\mb{I}^{d\times d}$ is a $d$-dimensional identity matrix. 
The encoding function for the tail node is defined analogously. 
The score function for this approach is given by
\begin{align*}\label{eq:transd}
s(\langle u, r, v \rangle) = -\|\enc(u, \langle u, r, v \rangle) + &\mb{r} \\ - \enc(v, \langle u, r,& v \rangle)\|_2,
\end{align*}
where $\mb{r} \in \mathbb{R}^d$ is another trainable embedding parameter (one per relation). 
Finally, we use a standard max-margin loss with a single negative sample per positive edge:
\begin{equation}\label{eq:margin}
    L_{\textrm{edge}}(s(e), s(e^-)) = \max(0, 1 - s(e) + s(e)^-).
\end{equation}
\subsubsection*{Movielens-1M}
\vspace{-2mm}
Our second dataset is derived from the MovieLens-1M recommender system benchmark \cite{harper2016movielens}.
This is a standard recommender system benchmark, where the goal is to predict the rating that users assign movies.
However, unlike previous work, in our experiments we treat the user features (age, gender, and occupation) as sensitive attributes (rather than as additional feature information for the recommendation task). 
Following \citet{berg2017graph} we treat this recommendation task as an edge prediction problem between users and movies, viewing the different possible ratings as different edge relations. 

For this dataset we use a simple ``embedding-lookup'' encoder, where each user and movie is associated with a unique embedding vector in $\mathbb{R}^d$.
As a scoring function, we follow \citet{berg2017graph} and use a  log-likelihood approach:
\begin{equation*}
    s(\langle u, r, m \rangle) = \mb{z}_u^\top\mb{Q}_r\mb{z}_v-\log(\sum_{r' \in \mathcal{R}}\mb{z}_u^\top\mb{Q}_{r'}\mb{z}_v),
\end{equation*}
The relation matrices $\mb{Q}_r \in \mathbb{R}^{d \times d}$ are computed as:
\begin{equation*}
    \mb{Q}_r = a_{r, 1}\mb{P}_1 + a_{r,2}\mb{P}_2, 
\end{equation*}
where $a_{r, 1}, a_{r, 1} \in \R$ and $\mb{P}_1, \mb{P}_2 \in \R^{d \times d}$ are trainable parameters. 
In this case, the loss function is simply the negative of the log-likelihood score. 

\subsubsection*{Reddit}
\vspace{-2mm}
The final dataset we consider is based on the social media website Reddit---a popular, discussion-based website where users can post and comment on content in different topical communities, called ``subreddits''. 
For this dataset, we consider a traditional edge prediction task, where the goal is to predict interactions between users and subreddit communities. 

To construct the edge prediction task, we examined all comments from the month of November in $2017$, and we placed an edge between a user and a community if this user commented on that community at least once within this time period. 
We then took the $10$-core of this graph to remove low-degree nodes, which resulted in a graph with approximately $366$K users, $18$K communities, and $7$M edges.
Given this graph, the main task is to train an edge-prediction model on $90\%$ of the user-subreddit edges and then predict missing edges in a held-out test set of the remaining edges.  

Reddit is a pseudonymous website with no public user attributes.  
Thus, to define sensitive attributes, we treat certain subreddit nodes as {\em sensitive nodes}, and the sensitive attributes for users are whether or not they have an edge connecting to these sensitive nodes.
In other words, the fairness objective in this setting is to force the model to be invariant to whether or not a user commented on a particular community. 
To select the ``sensitive'' subreddit communities, we randomly sampled $10$ from the top-100 communities by degree.\footnote{We excluded the top-5 highest-degree outlying communities.} 
Note that this setting represents the extreme case where we want the model to be invariant with respect to the existence of particular edges in the input graph. 

As with MovieLens-1M, we use a simple ``embedding-lookup'' encoder. 
In this case, there is only a single relation type---indicating whether a Reddit user has commented on a ``subreddit'' community.
Thus, we employ a simple dot-product based scoring function,
   $s(\langle u, r, v \rangle) = \mb{z}_u^\top\mb{z}_v$,
and we use a max-margin loss as in Equation \eqref{eq:margin}. 

\subsection{Results}
We now address the core experimental questions (\textbf{Q1}-\textbf{Q3}). 

\begin{table*}[t]
\caption{Ability to predict sensitive attributes on the MovieLens data when using various embedding approaches. For gender attribute the score is AUC while for age and occupation attributes the score is micro averaged F1. The columns represent the different embedding approaches (e.g., with or without adversarial regularizatin) while the rows are the attribute being classified.}
\label{ml-table}
\begin{center}
\begin{small}
\begin{sc}
\begin{tabular}{lccccccr}
\toprule
MovieLens1M & \multicolumn{1}{p{1.5cm}}{\centering Baseline \\ No Adversary} & \multicolumn{1}{p{1.5cm}}{\centering Gender \\ Adversary} & \multicolumn{1}{p{1.5cm}}{\centering Age \\ Adversary} & \multicolumn{1}{p{1.5cm}}{\centering Occupation \\ Adversary}&\multicolumn{1}{p{1.5cm}}{\centering Comp. \\ Adversary} & \multicolumn{1}{p{1.5cm}}{\centering Majority \\ Classifier} & \multicolumn{1}{p{1.5cm}}{\centering Random \\ Classifier} \\
\midrule
Gender    & 0.712& 0.532& 0.541 & 0.551 & 0.511 & 0.5 & 0.5 \\
Age    & 0.412 & 0.341& 0.333 & 0.321 & 0.313 & 0.367 & 0.141\\
Occupation     & 0.146 & 0.141& 0.108 & 0.131 & 0.121 & 0.126 & 0.05 \\

\bottomrule
\end{tabular}
\end{sc}
\end{small}
\end{center}
\vskip -0.1in
\end{table*}

\begin{table}[t]
\caption{Ability to predict sensitive attributes on the Freebase15k-237 data when using various embedding approaches. AUC scores are reported, since all the sensitive attributes are binary. The mean rank on the main edge-prediction task is also reported.}
\label{fb15k-table}
\vskip 0.15in
\begin{center}
\begin{small}
\begin{sc}
\begin{tabular}{lcccr}
\toprule
FB15k-237 & \multicolumn{1}{p{1.5cm}}{\centering Baseline \\ No Adversary} & \multicolumn{1}{p{1.5cm}}{\centering Non \\ Comp. Adversary} & \multicolumn{1}{p{1.5cm}}{\centering Comp. \\ Adversary} \\
\midrule
Attribute 0    &0.97& 0.82& 0.77  \\
Attribute 1 & 0.99& 0.81& 0.79\\
Attribute 2   & 0.98&  0.81 & 0.81 \\
Mean Rank &285& 320 & 542 \\
\bottomrule
\end{tabular}
\end{sc}
\end{small}
\end{center}
\vskip -0.1in
\label{fb_exp}
\end{table}

\begin{figure}[t!]
    \centering
    \includegraphics[width=0.9\linewidth]{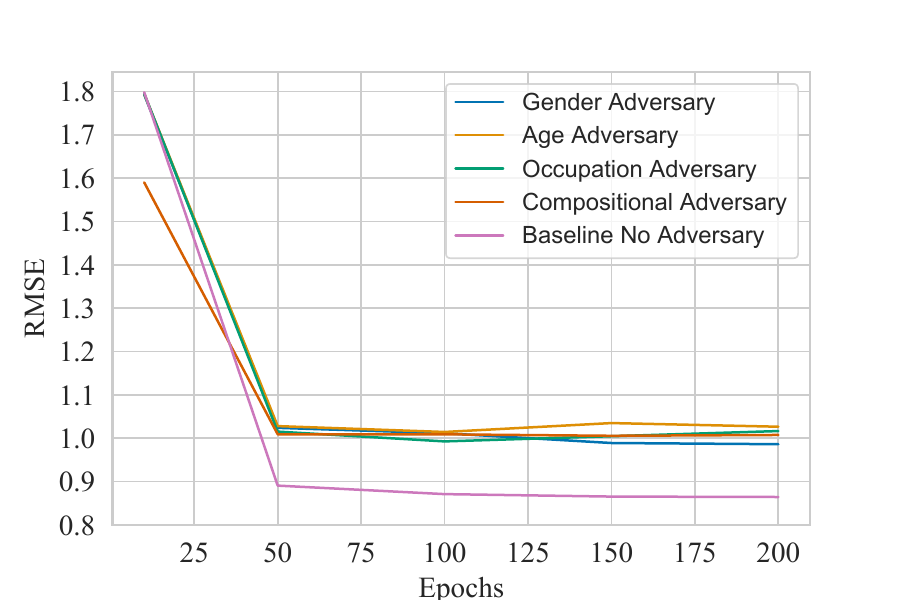}
    \vspace{-15pt}
    \caption{Performance on the edge prediction (i.e., recommendation) task on MovieLens, using RMSE as in \citet{berg2017graph}.}
    \label{fig:rmse}
    \vspace{-1mm}
\end{figure}
\begin{figure}[t!]
    \centering
    \includegraphics[width=0.9\linewidth]{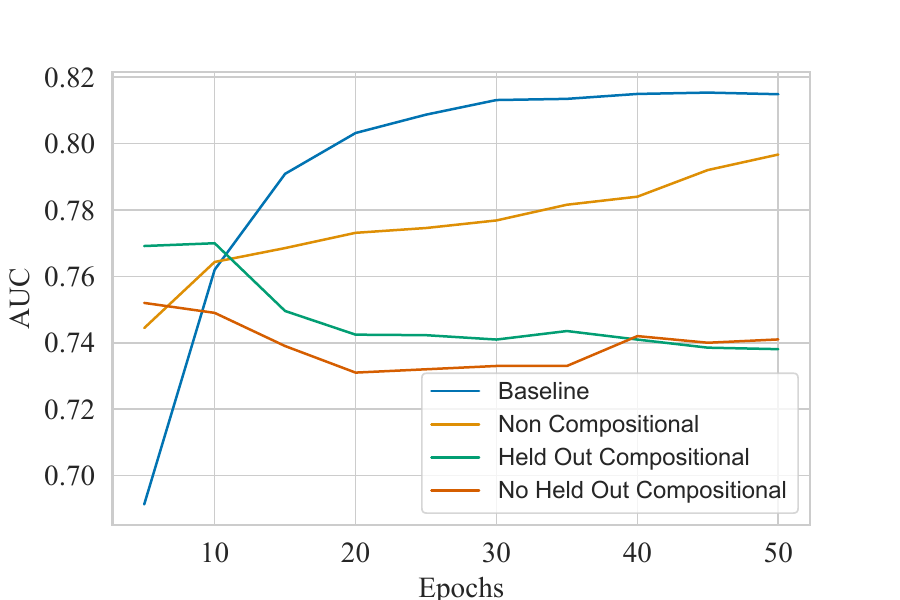}
       \vspace{-10pt}
    \caption{Performance on the edge prediction (i.e., recommendation) task on the Reddit data. Evaluation is using the AUC score, since there is only one edge/relation type.}
    \label{fig:reddit_enc_auc}
    \vspace{-2mm}
\end{figure}
\begin{figure}[t!]
    \centering
    \includegraphics[width=0.9\linewidth]{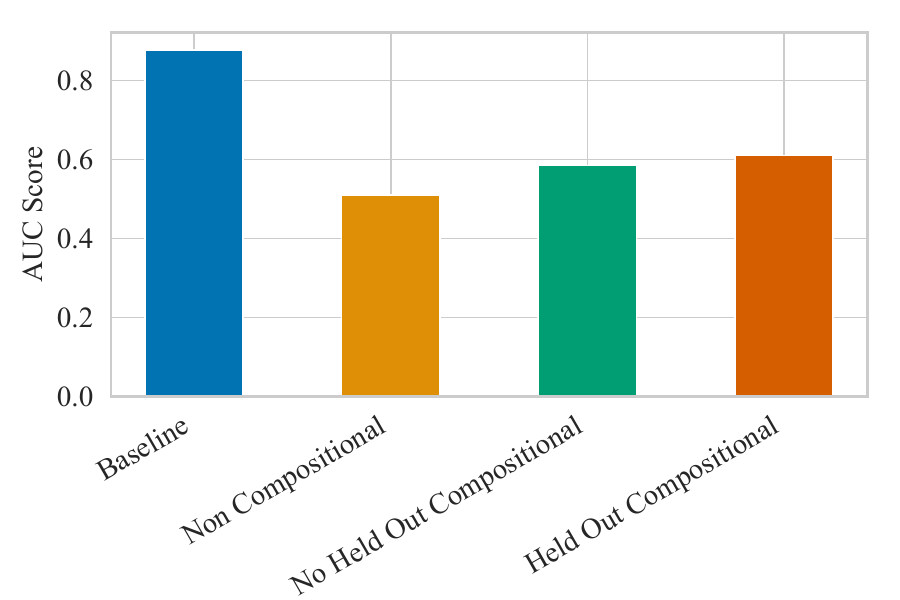}
    \vspace{-15pt}
    \caption{Ability to predict sensitive attributes on the Reddit data when using various embedding approaches. Bar plots correspond to the average AUC across the $10$ binary sensitive attributes.}
    \label{fig:reddit_auc}
    \vspace{-2mm}
\end{figure}

\subsubsection*{Q1: The Invariance-Accuracy Tradeoff}

In order to quantify the extent to which the learned embeddings are invariant to the sensitive attributes (e.g., after adversarial training), we freeze the trained compositional encoder $\compenc$ and train an new MLP classifier to predict each sensitive attribute from the filtered embeddings (i.e., we train one new classifier per sensitive attribute).
We also evaluate the performance of these filtered embeddings on the original prediction tasks. 
In the best case, a newly trained MLP classifier should have random accuracy when attempting to predict the sensitive attributes from the filtered embeddings, but these embeddings should still provide strong performance on the main edge prediction task. Thus, for binary sensitive attributes, an ideal result is an AUC score of $0.5$ when attempting to predict the sensitive attributes from the learned embeddings. 

\begin{figure}
    \centering
    \includegraphics[width=0.9\linewidth]{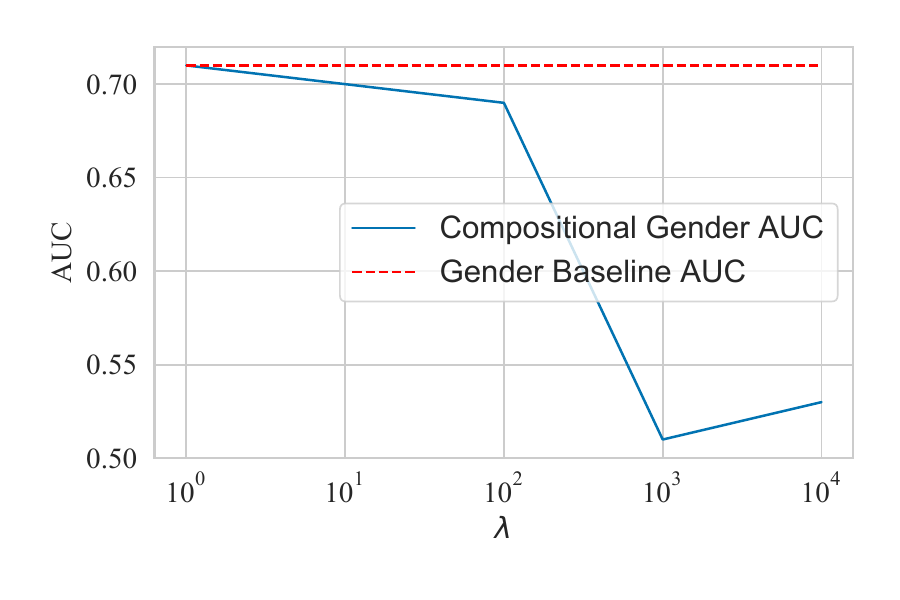}
      \vspace{-15pt}
    \caption{Tradeoff of Gender AUC score on MovieLens1M for a compositional adversary versus different $\lambda$}
    \label{fig:comp_gender_auc}
\vspace{-2mm}
\end{figure}

\begin{figure}
    \centering
    \includegraphics[width=0.9\linewidth]{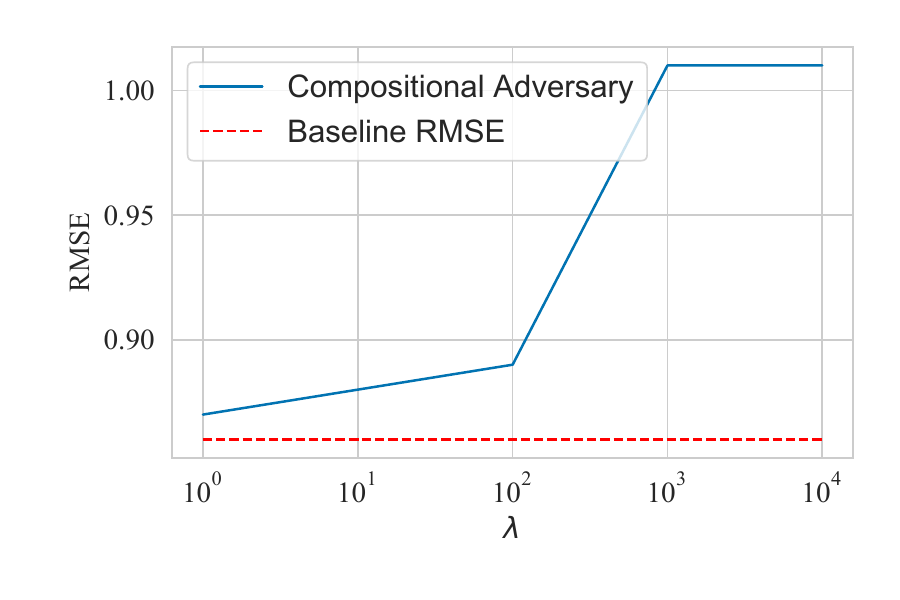}
      \vspace{-10pt}
    \caption{RMSE on MoveLens1M with various $\lambda$.}
    \label{fig:comp_gender_rmse}
\end{figure}

Overall, we found that on the more realistic social recommendation datasets---i.e., the MovieLens-1M and Reddit datasets---our approach was able to achieve a reasonable tradeoff, with the near-complete removal of the sensitive information leading to a roughly 10\% relative error increase on the edge prediction tasks. In other words, on these two datasets the sensitive attributes were nearly impossible to predict from the filtered embeddings, while the accuracy on the main edge prediction task was roughly 10\% worse than a baseline approach that does not include the invariance constraints.
Table \ref{ml-table} and Figure \ref{fig:rmse} summarize these results for the MovieLens data, where we can see that the accuracy of classifying the sensitive attributes is on-par with a majority-vote classifier (Table \ref{ml-table}) while the RMSE degrades from $0.865$ to $1.01$ with the compositional adversary. Figures \ref{fig:comp_gender_auc} and \ref{fig:comp_gender_rmse} illustrate this tradeoff and show how the RMSE for the edge prediction task and ability to predict the sensitive attributes change as we vary the regularization strength, $\lambda$. As expected, increasing $\lambda$ does indeed produce more invariant embeddings but leads to higher RMSE values.
Figures \ref{fig:reddit_enc_auc} and \ref{fig:reddit_auc} similiarly summarize these results on Reddit. 

Interestingly, we found that on the Freebase15k-237 dataset it was not possible to completely remove the sensitive information without incurring a significant decrease in accuracy on the original edge prediction task. \cut{As we can see, for all values of $\lambda$ the ability to predict the sensitive attributes remains high (i.e., above 0.75 AUC) while the mean rank begins to degrade with higher $\lambda$ values.}
This result is not entirely surprising, since for this dataset the ``sensitive'' attributes were synthetically constructed from entity type annotations, which are presumably very relevant to the main edge/relation prediction task. 
However, it is an interesting point of reference that demonstrates the potential limitations of removing sensitive information from learned graph embeddings.  

\subsubsection*{Q2: The Impact of Compositionality}

In all our experiments, we observed that our compositional approach performed favorably compared to an approach that individually enforced fairness on each individual attribute. In fact, on the MovieLens-1M data (and the synthetic Freebase15k-237 data), the compostionally trained adversary {\em outperformed} the individually trained adversaries in terms of removing information about the sensitive attributes (Table \ref{ml-table}). 
In other words, training a model to jointly remove information about the sensitive attributes using the compositional encoder (Equation \ref{eq:compenc}) removed more information about the sensitive attributes than training separate adversarially regularized embedding models for each sensitive attribute.
This result is not entirely surprising, as it essentially indicates that the different sensitive attributes (age, gender, and occupation) are correlated in this dataset. 
Nonetheless, it is a positive result indicating that the extra flexibility afforded by the compositional approach does not necessarily lead to a decrease in performance. 
That said, on the Reddit data we observed the opposite trend and found that the compositional approach performed worse in terms of its ability to remove information about the sensitive attributes (Figure \ref{fig:reddit_auc}) as well as a small drop on the performance of the main edge prediction task (Figure \ref{fig:reddit_enc_auc}). 

\subsubsection*{Q3: Invariance on Unseen Combinations}\label{subsubsec:comp_generalizability}

One of the key benefits of the compositional encoder is that it can flexibly generate embeddings that are invariant to any subset of $S \subseteq \{1, ..., K\}$ of the sensitive attributes in a domain. 
In other words, at inference time, it is possible to generate $2^K$ distinct embeddings for an individual node, depending on the exact set of invariance constraints. 
However, given this combinatorially large output space, a natural question is whether this approach performs well when generalizing to unseen combinations of sensitive attributes.

We tested this phenomenon on the Reddit dataset, since it has the largest number of sensitive attributes (10, compared to 3 sensitive attributes for the other two datasets). 
During training we held out $10\%$ of the combinations of sensitive attributes, and we then evaluated the model's ability to enforce invariance on this held-out set.
As we can see in Figure \ref{fig:reddit_auc}, the performance drop for the held-out combinations is very small ($0.025$), indicating that our compositional approach is capable of effectively generalizing to unseen combinations. 
The Appendix contains further results demonstrating how this trends scales gracefully when we increase the number of sensitive attributes from 10 to 50. 

\subsubsection*{Quantifying Bias}
In all of the above results, we used the ability to classify the sensitive attributes as a proxy for bias being contained within the embeddings.
While this is a standard approach, e.g., see \citet{edwards2015censoring}, and an intuitive method for evaluating representational invariance---a natural question is whether the adversarial regularization also decreases bias in the edge prediction tasks. 
Ideally, after filtering the embeddings, we would have that the edge predictions themselves are not biased according to the sensitive attributes.

\begin{figure}
    \centering
    \includegraphics[width=0.9\linewidth]{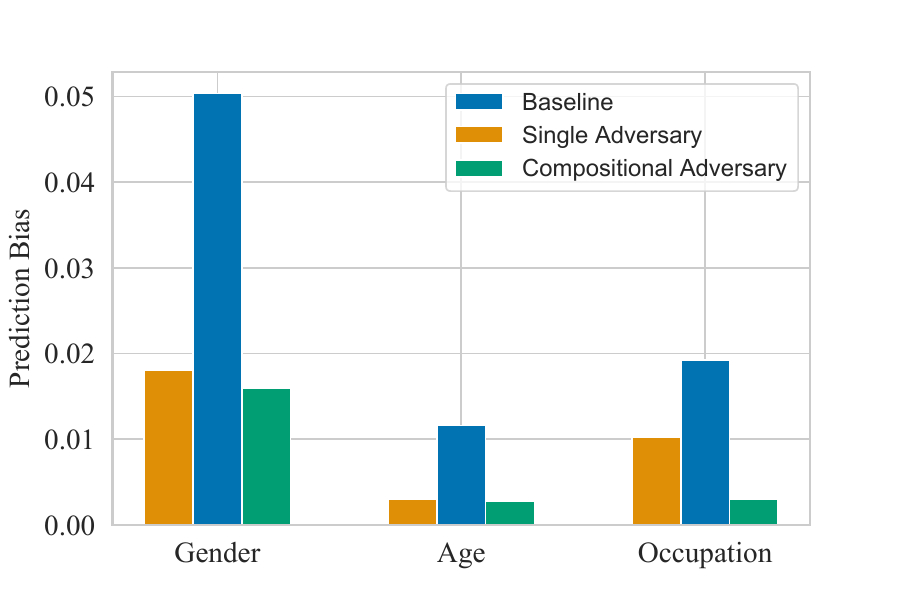}
    \vspace{-15pt}
    \caption{Prediction Bias for different Sensitive Attributes under three settings in MovieLens1M.}
    \label{fig:pred_bias}
    \vspace{-2.5mm}
\end{figure}

To quantify this issue, we computed a ``prediction bias'' score for the MovieLens1M dataset: For each movie, we computed the absolute difference between the average rating predicted for each possible value of a sensitive attribute and we then averaged these scores over all movies. 
Thus, for example, the bias score for gender corresponds to the average absolute difference in predicted ratings for male vs.\@ female users, across all movies. From the perspective of fairness our adversary imposes a soft demographic parity constraint on the main task. A reduction in prediction bias across the different subgroups \cut{for a sensitive attribute} represents an empirical measure of achieving demographic parity.
Figure \ref{fig:pred_bias} highlights these results, which show that adversarial regularization does indeed drastically reduce prediction bias. Interestingly, using a compositional adversary works better than a single adversary for a specific sensitive attribute which we hypothesize is due to correlation between sensitive attributes.

\vspace{-2mm}
\section{Discussion and Conclusion}
\cut{
We introduced an adversarial framework to enforce fairness constraints on graph embeddings. We present two different strategies two enforce fairness: (1) a non-compositional adversary and (2) a compositional adversary. Each type of adversary critiques the representations learned by the main model such that sensitive attributes are not encoded by the main model.
In the compositional approach, we learn a bank of adversarial filters to optionally enforce fairness constraints over a set of possible sensitive attributes.
}

Our work sheds light on how fairness can be enforced in graph representation learning---a setting that is highly relevant to large-scale social recommendation and networking platforms. 
We found that using our proposed compositional adversary allows us to flexibly accomodate unseen combinations of fairness constraints without explicitly training on them. This highlights how fairness could be deployed in a real-word, user-driven setting, where it is necessary to optionally enforce a large number of possible invariance constraints over learned graph representations. 

In terms of limitations and directions for future work, one important limitation is that we only consider one type of adversarial loss to enforce fairness.
While this adversarial loss is theoretically motivated and known to perform well, there are other recent variations in the literature (e.g., \citet{madras2018learning})---as well as related non-adversarial regularizers (e.g., \citet{zemel2013learning}). Also, while we considered imposing fairness over sets of attributes, we did not explicitly model subgroup-level fairness \cite{kearns2017preventing}.
Extending and testing our framework with these alternatives is a natural direction for future work. 

There are also important questions about how our framework translates to real-world production systems. For instance, in this work we enforced fairness with respect to randomly sampled sets of attributes, but in real-world environments, these sets of attributes would come from user preferences, which may themselves be biased; e.g., it might be more common for female users to request fairness than male users potentially leading to new kinds of demographic inequalities. Understanding how these preference biases could impact our framework is an important direction for future inquiry.

\subsection*{Acknowledgements}
The authors would like to thank the anonymous ICML reviewers for their helpful comments. In addition we would like to thank Koustuv Sinha, Riashat Islam and Andre Cianflone for helpful feedback on earlier drafts of this work.
The researchers also thank Jason Baumgartner for his creation and curation of the pushshift.io Reddit data. 
This research was funded in part by an academic grant from Microsoft Research, as well as a Canada CIFAR Chair in AI, held by Prof.\@ Hamilton.
\bibliography{bibliography.bib}
\bibliographystyle{icml2019}

\clearpage
\appendix
\section{Implementation Details}
We implement each discriminator and adversarial filter as  multi-layer perceptrons (MLPs) with a leaky ReLU non-linearity between layers, and we use the Adam optimizer with default parameters. Unless otherwise specified we use $\lambda=1000$ for all experiments and datasets. For fair comparison our discriminator during training time and subsequent sensitive attribute classifier share the same architecture and capacity. Finally, for every step performed by the main encoding model the Discriminator is updated $5$ times. We found that this was necessary to provide a sufficient supervisory signal to the main encoding model.

\section{FB15k-237 Details}
To generate negative triplets we randomly sample either a head or tail entity during training, with a ratio of $20$ negatives for each positive triplet. The TransD model is trained for $100$ epochs with an embedding dimension of $20$, selected using cross-validation, while the sensitive attribute classifers are trained for $50$ epochs. The discriminators, sensitive attribute classifier and adversarial filters are modelled as MLP's with $4$,$4$ and $2$ layers respectively. Lastly, we use the training, validation and testing splits provided in the datasets.

\section{MovieLens1M}
As with FB15k-237 we use model the discriminators and sensitive attribute classifiers are modelled as MLP's but $9$ layers with dropout with $p=0.3$ between layers while the adversarial filter remains unchanged from FB15k-237. We found that regularization was crucial to the performance of main model and we use BatchNorm after the embedding lookup in the main model which has an embedding dimensionality of $30$. As only user nodes contain sensitive attributes our discriminators do not compute losses using movie nodes. Finally, to train our sensitive attribute classifier we construct a $90\%$ split of all users while the remaining user nodes are used for test. The same ratio of train/test is used for the actual dataset which constains users,movies and corresponding ratings for said movies. Finally, we train the main model and sensitive attribute classifiers for $200$ epochs.

\section{Reddit}
Like FB15k-237 we generate negative triplets by either sampling head or tail entities which are either users or subreddits but unnlike FB15k-237 we keep the ratio of negatives and positives the same. We also inherit the same architectures for discriminator, sensitive attribute classifier and attribute filters used in MovieLens1M. The main model however uses an embedding dimensionality of $50$. Similar to MovieLens1M only user nodes contain sensitive attributes and as such the discriminator and sensitive attribute classifier does not compute losses with respect to subreddit nodes. Also, our training set comprises of a $90\%$ split of all edges while the the remaining $10\%$ is used as a test set. To test compositional generalizability we held out $10\%$ of user nodes. Lastly, we train the main model for $50$ epochs and the sensitive attribute classifier for $100$ epochs.
\section{Additional Results on Reddit}
To the test degree of which invariance is affected by the number of sensitive attributes we report additional results on the Reddit dataset. Specifically, we report results for the Held out set with  20, 30, 40, and 50 sensitive attributes. Overall, these results show no statistically significant degradation in terms of invariance performance or task accuracy. 

\begin{table}[t]
\caption{Average AUC values across top-k sensitive attributes for Reddit. The results are reported on a Held Out test of different combinations of attributes.}
\label{sample-table}
\vskip 0.15in
\begin{center}
\begin{small}
\begin{sc}
\begin{tabular}{lcr}
\toprule
Reddit & \multicolumn{1}{p{1.5cm}}{\centering Held out AUC} \\
\midrule
20 Sensitive Attributes  & 0.569 \\
30 Sensitive Attributes  & 0.569 \\
40 Sensitive Attributes  & 0.556  \\
50 Sensitive Attributes  & 0.519 \\

\bottomrule
\end{tabular}
\end{sc}
\end{small}
\end{center}
\vskip -0.1in
\label{fb_exp}
\end{table}

\end{document}